\documentclass[final]{cvpr}

\usepackage{times}
\usepackage{epsfig}
\usepackage{graphicx}
\usepackage{amsmath}
\usepackage{amssymb}
\usepackage{color}
\usepackage{times}
\usepackage{epsfig}
\usepackage{algorithm}
\usepackage{algorithmicx}
\usepackage{algpseudocode}
\usepackage{graphics}
\usepackage{threeparttable}
\usepackage{color}
\usepackage[normalem]{ulem}
\usepackage{multirow}
\usepackage{float}
\usepackage{amsfonts}
\usepackage{bm}
\usepackage{subfig}
\usepackage{array}
\usepackage[table]{xcolor}
\usepackage{colortbl}
\usepackage{pifont}
\usepackage{cite}
\usepackage{diagbox}
\usepackage{rotating}
\usepackage{booktabs}
\usepackage{overpic}
\usepackage{textcomp}
\usepackage{contour}
\usepackage{enumitem}

\usepackage{etoolbox}
\usepackage[marginal]{footmisc}
\usepackage[utf8]{inputenc}

\definecolor{mygray}{gray}{.92}

\newcommand{\tabincell}[2]{\begin{tabular}{@{}#1@{}}#2\end{tabular}}

\usepackage[pagebackref=true,breaklinks=true,letterpaper=true,colorlinks,bookmarks=false]{hyperref}
\usepackage{cleveref}
\crefname{section}{§}{§§} 
\Crefname{section}{§}{§§}

\def\cvprPaperID{} 

\usepackage[pagebackref=true,breaklinks=true,letterpaper=true,colorlinks,bookmarks=false]{hyperref}
\usepackage{colortbl}
\definecolor{mygray}{gray}{.9} 
\def\cvprPaperID{1111} 

\def\cvprPaperID{1178} 
\def\confYear{CVPR 2021}

\begin{document}
\title{Mutual Graph Learning for Camouflaged Object Detection}

\author{
Qiang Zhai$^{1,\ddag}$~
Xin Li$^{2,\ddag}$~
Fan Yang$^{2,}$\thanks{Corresponding author: Fan Yang \emph{(fanyang\_uestc@hotmail.com)}}~~
Chenglizhao Chen$^{3}$~
Hong Cheng$^{1}$~ 
Deng-Ping Fan$^4$\\
$^1$ UESTC \quad
$^2$ Group 42 (G42) \quad
$^3$ Qingdao University \quad
$^4$ Inception Institute of AI (IIAI)\\
{\tt \small $\ddag$ Equal contributions 
}
\\
}

\maketitle
\pagestyle{empty}
\thispagestyle{empty}

\begin{abstract}
	Automatically detecting/segmenting object(s) that blend in with their surroundings is difficult for current models. A major challenge is that the intrinsic similarities between such foreground objects and background surroundings make the features extracted by deep model indistinguishable. To overcome this challenge, an ideal model should be able to seek valuable, extra clues from the given scene and incorporate them into a joint learning framework for representation co-enhancement. With this inspiration, we design a novel Mutual Graph Learning ({MGL}) model, which generalizes the idea of conventional mutual learning from regular grids to the graph domain. Specifically, MGL decouples an image into two task-specific feature maps --- one for roughly locating the target and the other for accurately capturing its boundary details --- and fully exploits the mutual benefits by recurrently reasoning their high-order relations through graphs. Importantly, in contrast to most mutual learning approaches that use a shared function to model all between-task interactions, MGL is equipped with typed functions for handling different complementary relations to maximize information interactions. Experiments on challenging datasets, including CHAMELEON, CAMO and COD10K, demonstrate the effectiveness of our {MGL} with superior performance to existing state-of-the-art methods. Code is available at \url{https://github.com/fanyang587/MGL}.

\end{abstract}

\section{Introduction}
Camouflage is an important skill in nature, because it helps certain animals hide from their predators by blending in with their surroundings. The ability of camouflaging, which is closely related to how human perception works, has attracted increasing research attention over past decades. Biological and psychological studies show that it is hard for human beings to quickly spot camouflaged animals or objects~\cite{stevens2009animal,cuthill2019camouflage}. A possible reason is that the primitive function of our visual system may be designed to detect topological properties~\cite{chen1982topological}, thus making it difficult to identity camouflaged animals/objects that 
break up visual edge information of their `true' bodies. In spite of these biology discoveries, how to make up for this `flaw' in human perception by \texttt{Machine} is, unfortunately, still an under-explored topic in computer vision. 

\begin{figure}[pt]
	\begin{center}
		\includegraphics[width=0.99\linewidth]{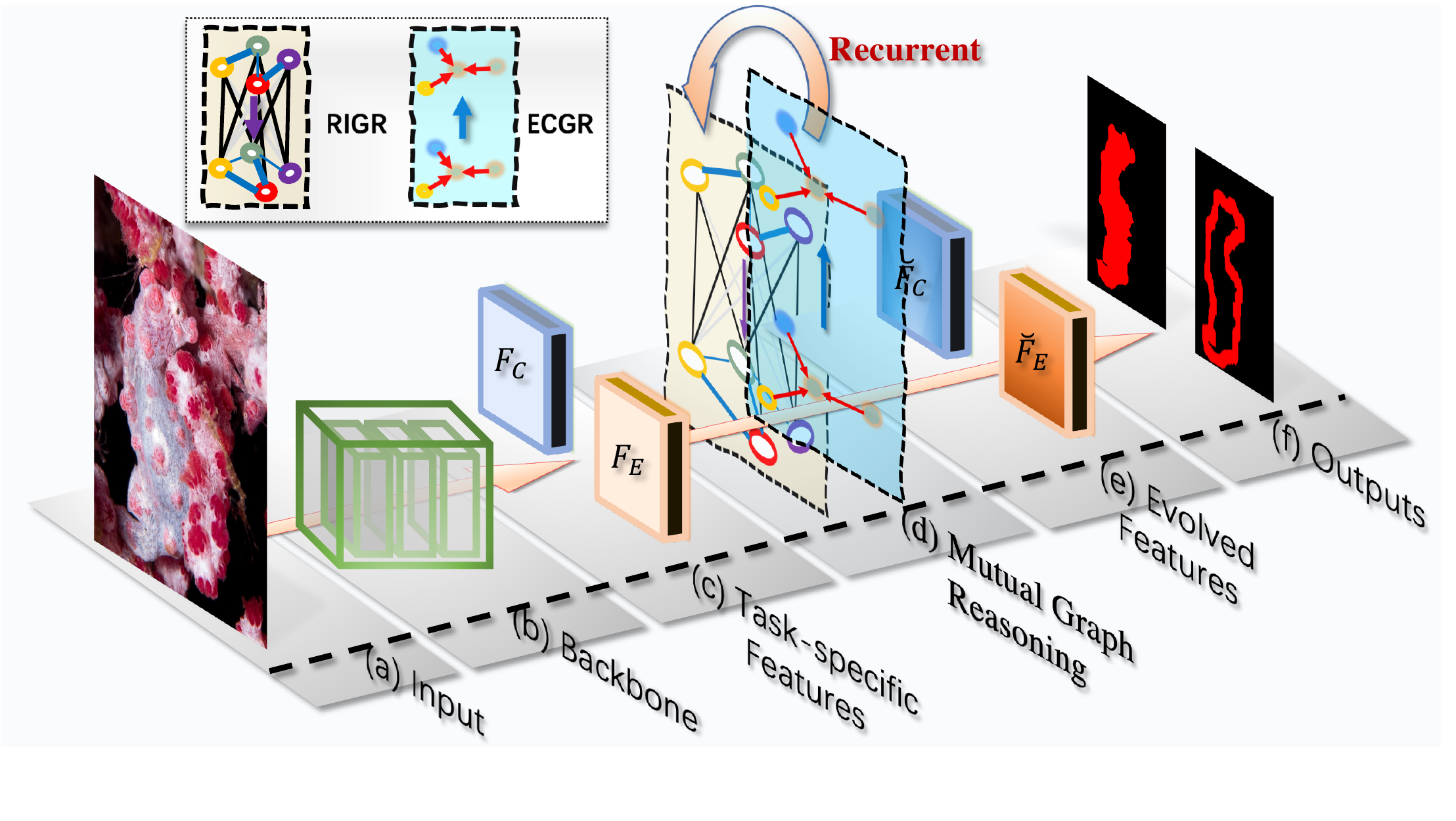}
	\end{center}
	\vspace{-25pt}
	\caption{{\bf Illustration of {MGL}}. Given an image (a), we use a ResNet-FCN as the backbone (b) to extract task-specific features for the camouflaged object detection (COD) and camouflaged object-aware edge extraction (COEE), respectively (c). Then, we exploit the mutual benefits from both tasks by reasoning about their mutual relations with the cooperation of the Region-Induced Graph Reasoning (RIGR) module and Edge-Constricted Graph Reasoning (ECGR) module in a recurrent manner (d). Finally, the evolved features (e) are mapped into the results (f).}
	\label{fig:1}
	\vspace{-1.5em}
\end{figure}

Identifying a camouflaged object from its background, also known as camouflaged object detection (COD)~\cite{fan2021cancealed}, is a valuable, yet challenging task~\cite{fan2020camouflaged}. `Seeing through camouflage' has promising prospects for facilitating various real-life tasks, including image retrieval~\cite{liu2013model}, species discovery~\cite{perez2012early}, traffic risk management, medical image analysis~\cite{fan2020inf,fan2020pranet,wu2021jcs}, \emph{etc}. However, the existing deep models are still incapable of fully resolving the intrinsic visual similarities between foreground objects and background surroundings. To overcome this difficulty, current approaches distill additional knowledge by extracting auxiliary features from the shared context, \emph{e.g.,} features for identification~\cite{fan2020camouflaged} or classification~\cite{ltnghia-CVIU2019}, to significantly augment the underlying representations for camouflaged object detection. Although their notable successes truly demonstrate the benefit of exploiting extra knowledge in camouflaged object detection, there are still three major open issues. \textbf{First}, the mutual influence between COD and its auxiliary task is overlooked or poorly investigated. More specifically, because the existing efforts~\cite{ltnghia-CVIU2019,fan2020camouflaged,yan2020} only exploit extra information from the auxiliary task to guide/assist the main task (\emph{i.e.} COD), while ignoring the important collaborative relationship between them, these models may fail to a local minimum~\cite{szegedy2016rethinking}. \textbf{Second}, as the cross-task dependencies are modeled only in the original coordinate space, more global, higher-order guidance information may be lost. As we demonstrate empirically, current COD models become ineffective under heavy occlusions and indefinable boundaries, because they fail to incorporate higher-order information into the representation learning process. \textbf{Third}, according to recent biological discoveries~\cite{webster2013disruptive,webster2015does,kang2015camouflage}, a key factor for concealment/camouflage is the \emph{edge disruption}. Unfortunately, how to enhance true edge visibility for facilitating the representation learning for COD is not investigated by existing arts~\cite{ltnghia-CVIU2019,fan2020camouflaged}, which definitely would weaken, or at least not fully utilize, the COD model's learning power.

Targeting at these drawbacks, we present a novel Mutual Graph Learning model ({MGL}) to sufficiently and comprehensively exploit mutual benefits between camouflaged object detection (COD) and its auxiliary task. Considering that the \emph{edge disruption} should be one of the key factors for camouflage~\cite{webster2013disruptive,webster2015does,kang2015camouflage}, we treat the camouflaged object-aware edge extraction (COEE) as an auxiliary task and incorporate it into our {MGL} for mutual learning. As shown in Figure~\ref{fig:1}, our MGL has a well-designed interweaving architecture that strengthens the interaction and cooperation between tasks. Importantly, instead of `na\"ively' fusing the learned features from two tasks as in the existing works, MGL precisely exploits useful information from the counterparts for representation co-enhancement by explicitly reasoning about the complementary relations between COD and COEE with two typed functions. To mine the semantic guidance information from COD and assist COEE, we develop a novel Region-Induced Graph Reasoning (RIGR) module to reason about the high-level dependencies, and transfer semantic information from COD to augment underlying representations for COEE; To improve the true edge visibility, a new Edge-Constricted Graph Reasoning (ECGR) module is used to explicitly incorporate the edge information from COEE to, in turn, better guide the representation learning for COD. Importantly, our RIGR and ECGR can be formulated in a recurrent manner to recursively mine the mutual benefits and incorporate valuable information from their counterparts.

We demonstrate the effectiveness of our {MGL} by comparing it against strong baselines and current state-of-the-art methods through extensive experiments on a variety of benchmarks. The experiment results clearly demonstrate its superiority over existing methods in mining mutual guidance information for camouflaged object detection. The contributions of this work are summarized as follows:

\begin{itemize}[leftmargin=*]
	\vspace{-0.75em}
	\item{\bf  A novel graph-based, mutual learning approach for camouflaged object detection.} To our knowledge, this is the first attempt to exploit mutual guidance knowledge between two closely related tasks, \emph{i.e.,} COD and COEE, using the graph-based techniques for camouflaged object detection. This approach is able to capture semantic guidance knowledge and spatial supportive information for mutually boosting the performance of both tasks. 
	
	\vspace{-0.75em}
	\item{\bf  Carefully designed graph-based interaction functions for fully mining typed guidance information.} Unlike conventional mutual learning approaches, our {MGL} ensembles two distinct graph-based interaction modules to reason about typed relations: {\bf RIGR} for mining semantic guidance information from COE to assist COEE and, {\bf ECGR} for incorporating true edge priors to enhance the underlying representations of COD. 
	
	
	\vspace{-0.75em}
	\item{\bf State-of-the-art results on widely-used benchmarks.} Our {MGL} sets new records on a variety of benchmarks, \emph{i.e.,} \emph{CHAMELEON}~\cite{skurowski2018animal}, \emph{CAMO}~\cite{ltnghia-CVIU2019} and \emph{COD10K}~\cite{fan2020camouflaged}, and outperforms existing COD models by a large margin. 
	
\end{itemize}

\section{Related Work}
\noindent {\bf Camouflaged Object Detection.}  The camouflaged object detection (COD) task~\cite{mei2021Ming,aixuan_cod_sod21,zhai2021Mutual} has posed new challenges by pushing the boundaries of generic / salient object detection~\cite{redmon2016you,wei2020label,li2016saliency,luo2020webly,he2017mask,li2017multi,lin2017feature,liu2016ssd,lin2017focal,zhuge2021salient,zhao2019egnet,GateNet,Li_2018_ECCV,pang2020multi} to concealed objects blending in with their surroundings. Fan~\emph{et~al.}~\cite{fan2020camouflaged}  present the Search and Identification Net (SINet) to address this challenge by first roughly searching for camouflaged objects and then performing segmentation. Le~\emph{et~al.}~\cite{ltnghia-CVIU2019} introduce the Anabranch Network (ANet) which incorporates classification information into representation learning. Yan~\emph{et~al.}~\cite{yan2020} introduce MirrorNet to use both instance segmentation and adversarial attack for COD. The common idea behind these bio-inspired models is that exploring and integrating extra clues into representation learning can greatly outperform the conventional approaches for generic object detection (GOD) and salient object detection (SOD) ~\cite{zhou2020interactive,lin2017feature,he2017mask,zhao2017pyramid,liu2018picanet,huang2019mask,qin2019basnet,qin2021boundary,deng2021re}. Unlike prior works, our novelty is that we use a unified, graph-based model to simultaneously perform camouflaged object detection (COD) and the camouflaged object-aware edge extraction (COEE) by comprehensively reasoning about multi-level relations to boost performance for both tasks.

\begin{figure*}[pt]
	\begin{center}
		\includegraphics[width=0.999\linewidth]{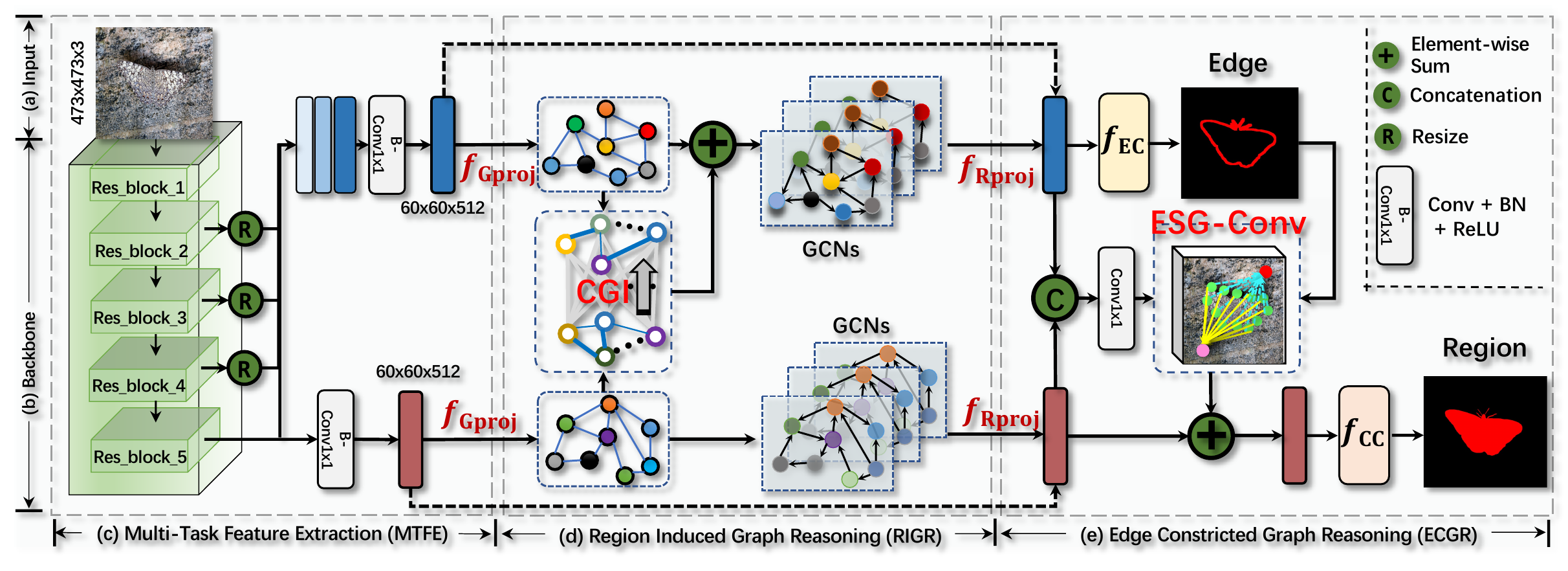}
	\end{center}
	\vspace{-20pt}
	\caption{{\bf An overview of our proposed (single-stage) mutual graph learning framework ({S-MGL}}). The main components of the flowchart are marked from (a) to (e). {\bf CGI} means the cross-graph interaction module and {\bf ESG-Conv} means the edge supportive graph convolution, which are the key operations for information interactions. Please refer to \cref{sec:ourmodel} for details.}
	\label{fig:2}
	\vspace{-1.5em}
\end{figure*}

\noindent {\bf Graph Convolutional Networks.} GCNs are powerful tools for graph data analysis, which have given rise to many applications~\cite{Yang_2020_CVPR,Wei_2020_CVPR,Xie-SAGNN,luo2020hybrid,Mohamed_2020_CVPR,Zhang_2020_CVPR, xie2020region}. In the context of (generic/salient) object detection, GCNs are used to detect or segment 2D/3D objects in images, videos or point clouds~\cite{wang2019dynamic,wang2019graph}. In~\cite{Chen_2019_CVPR,li2020spatial}, the long-range context is modeled by graph convolution for semantic segmentation. Wu~\emph{et~al.}~\cite{Wu_2020_CVPR} exploit the semantic relations and co-occurrence among objects and background with a bidirectional graph. Luo~\emph{et~al.}~\cite{Luo2020CascadeGN} introduce a cascade graph model to exploit multi-scale, cross-modality information for salient object detection. In~\cite{Zhang_2020_CVPR}, an adaptive GCN model with attention graph clustering is introduced for co-saliency detection. For camouflaged object detection, we introduce two novel graph-based modules, RIGR and ECGR, to fully reason about complementary information of COD and COEE across different levels, which can better learn representations from image to overcome multiple challenges.

\section{Our Approach}\label{sec:ourmodel}
\subsection{Preliminaries}
\noindent {\bf Motivation.} Our method is inspired by the discoveries from biological research~\cite{webster2013disruptive,webster2015does,kang2015camouflage}: capturing the true body/object shape is the key to seeing through camouflage. Then, an ideal model for camouflaged object detection should be well capable of capturing true edges of objects and, more importantly, incorporating such information into a joint learning framework. Intuitively, the involved tasks can benefit each other by information propagation in a unified, graph-based network.

\noindent {\bf Problem Formulation.}
Let the COD model be represented by the function {$\mathcal M_{\Theta}$} parameterized by weights {$\Theta$}, that takes an image $\mathbf I$ as input, and produces camouflage map $\mathbf C \in [0,1]$ and camouflaged object-aware edge map $\mathbf E \in [0,1]$ simultaneously, which reflect the probability of each pixel belonging to the camouflaged object(s) and its edges respectively. Our goal is to learn {$\Theta$} by fully exploiting the mutual benefits between COD and COEE, given the labeled training dataset $\{I_i, C_i, E_i^t\}_{i=1}^N$, where $I_i$ is a training image, $C_i$ means its groundtruth camouflage map, and $E_i$ denotes the true edge map which can be automatically generated from $C_i$.


\subsection{Overview}
MGL consists of three major components: Multi-Task Feature Extraction ({\bf MTFE}), Region-Induced Graph Reasoning ({\bf RIGR}) module and Edge-Constricted Graph Reasoning ({\bf ECGR}).

\begin{itemize}[leftmargin=*]
	\vspace{-0.6em}
	\item{\bf MTFE.} Given an input image $\mathbf I\in \mathbb{\bm R}^{H\times W\times 3}$, a multi-task backbone network $f_{\tt MTFE}$ decouples it into two task-specific representations: $\mathbf F_C\in \mathbb{\bm R}^{h\times w\times c}$ for roughly detecting the target and $\mathbf F_E\in \mathbb{\bm R}^{h\times w\times c}$ for properly capturing its true edges.
	
	\vspace{-0.6em}
	\item{\bf RIGR.} In this stage, $\mathbf F_C$ and $\mathbf F_E$ are first transformed into sample-dependent semantic graphs $\mathcal G_C= (\mathcal V_C, \mathcal E_C)$ and $\mathcal G_E = (\mathcal V_E, \mathcal E_E)$ by the graph projection operation $f_{\tt Gproj}$, where pixels with similar features form a vertex and edges measure the affinity between vertices in a feature space. Then, Cross-Graph Interaction module ({CGI}) $f_{\tt CGI}$ is used to capture the high-level dependencies between $\mathcal G_C$ and $\mathcal G_E$ and transfer semantic information from $\mathcal V_C$ to $\mathcal V_E$: $\mathcal V'_E = f_{\tt CGI} (\mathcal V_C, \mathcal V_E)$. Next, graph reasoning  $f_{\tt GR}$ is conducted to obtain evolved graph representations $\mathbf V_C$ and $\mathbf V'_E$ by graph convolution~\cite{kipf2016semi}. At last, $\mathbf V_C$ and $\mathbf V'_E$ are projected back to the original coordinate space $\hat{\mathbf F}_C = f_{\tt Rproj} (\mathbf V_C)$ and $\breve{\mathbf F}_E = f_{\tt Rproj} (\mathbf V'_E)$.

	\vspace{-0.6em}
	\item{\bf ECGR.} Before spatial relationship analysis, $\breve{\mathbf F}_E$ is first fed into the edge classifier $f_{\tt EC}$ to obtain camouflaged object-aware edge map $\mathbf E$. In addition, we fuse $\breve{\mathbf F}_E$ and $\hat{\mathbf F}_C$ (\emph{e.g.,} by {\tt concatenate}) to form a new feature map $\mathbf F'_C$ for COD, and then use a new Edge Supportive Graph Convolution ({\tt ESG-Conv}) to encode edge information and enhance $\mathbf F'_C$ for better locating objects, under the guidance of $\mathbf E$: $\breve{\mathbf F}_C = \mathtt{ESGConv} (\mathbf F'_C; \mathcal G^e(\mathbf E))$ where $\mathcal G^e(\mathbf E)$ denotes the edge supportive graph which is conditioned on $\mathbf E$. Finally, we feed $\breve{\mathbf F}_C$ into the classifier $f_{\tt CC}$ to obtain the final results $\mathbf C$. 
\end{itemize}

\noindent Figure~\ref{fig:2} presents an overview of our method. In {MGL}, the mutual relations between COD and COEE are reasoned over multiple levels of interaction spaces by employing two novel neural modules, \emph{i.e.,} RIGR and ECGR. By explicitly reasoning about their relationships, valuable mutual guidance information, intuitively, can be precisely propagated to assist each other during representation learning. It is worth mentioning that RIGR and ECGR can be stacked consecutively for recurrent mutual learning.

\subsection{Mutual Graph Learning}
Here, we give a detailed introduction to our Multi-Task Feature Extraction (MTFE), Region-Induced Graph Reasoning ({RIGR}) and Edge-Constricted Graph Reasoning ({ECGR}).

\noindent {\bf Multi-Task Feature Extraction (MTFE).} 
$f_{\tt MTFE}$ takes an image as the input, and produces two task-specific feature maps --- one for COD and the other for COEE. Formally, given an input image $\mathbf I\in \mathbb{\bm R}^{H\times W\times 3}$, a multi-task backbone network (\emph{i.e}, a multi-branch ResNet-based FCN network parameterized by $\Theta_{\tt MTFE}$) is employed to simultaneously obtain representations for COD ($\mathbf F_C$) and COEE ($\mathbf F_E$):
\begin{equation}  \textstyle
\begin{aligned}
\small
\mathbf F_C = f_{\tt MTFE} (\mathbf I; \Theta_{\tt MTFE}),~~~
\mathbf F_E = f_{\tt MTFE} (\mathbf I; \Theta_{\tt MTFE}),
\label{eq1}
\end{aligned}
\end{equation}
where $\mathbf F_C \in \mathbb{\bm R}^{h\times w\times c}$ and $\mathbf F_E \in \mathbb{\bm R}^{h\times w\times c}$ are features with $h\times w$ spatial resolution and $c$ channels for COD and COEE respectively, so that spatial information and high-level semantic information can be well preserved.

\begin{figure}[pt]
	\begin{center}
		\includegraphics[width=0.99\linewidth]{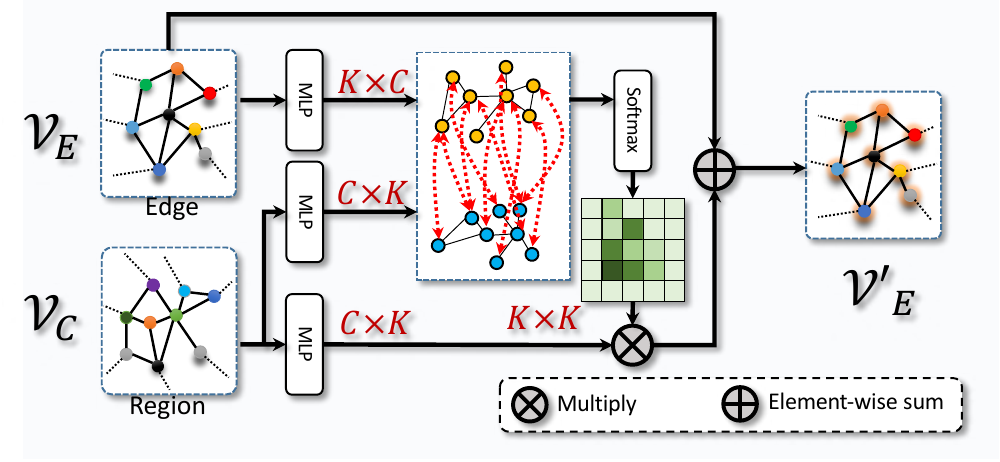}
	\end{center}
	\vspace{-20pt}
	\caption{{\bf Illustration of {CGI}}. CGI promotes the cross-graph (task) interaction, and transfers the information of COD to learn the evolved graph representations for COEE.}
	\label{fig:3}
	\vspace{-1.em}
\end{figure}

\noindent {\bf Region-Induced Graph Reasoning (RIGR).}
RIGR aims at reasoning about the region-induced semantic relations within COD and between COD and COEE, regardless of local details. It consists of four operations/functions: {\small \emph{(1)}} Graph Projection $f_{\tt Gproj}$, {\small \emph{(2)}} Cross-Graph Interaction $f_{\tt CGI}$, {\small \emph{(3)}} Graph Reasoning $f_{\tt GR}$ and {\small \emph{(4)}} Graph Reprojection $f_{\tt Rproj}$.

\noindent {\bf \small \emph{{(1)} Graph Projection}} 
$f_{\tt Gproj}$. Given input features $\mathbf F_C \in \mathbb{\bm R}^{h\times w\times c}$ or $\mathbf F_E \in \mathbb{\bm R}^{h\times w\times c}$, we first use a $1\times1$ convolutional layer to transform them into lower-dimension features, denoted as $\mathbf F^l_C \in \mathbb{\bm R}^{(h\times w)\times C}$ or $\mathbf F^l_E \in \mathbb{\bm R}^{(h\times w)\times C}$. Then, $f_{\tt Gproj}$ is used to transform feature vectors, $\mathbf F^l_C$ or $\mathbf F^l_E$, into graph node embeddings/representations, \emph{i.e.,} $\mathcal V_C \in \mathbb{\bm R}^{C\times K}$ or $\mathcal V_E \in \mathbb{\bm R}^{C\times K}$. Following~\cite{Zhang_2017_CVPR,li2018beyond}, we parameterize $f_{\tt Gproj}$ by $W \in \mathbb{\bm R}^{K\times C}$ and $\Sigma \in \mathbb{\bm R}^{K\times C}$. Each column $w_k$ of $W$ specifies a learnable clustering center for the $k$-th node. Specifically, the representation of each node can be computed as follow:
\begin{equation}  \textstyle
\begin{aligned}
\small
v_k = \frac{v'_k}{||v'_k||_2}, ~~ v'_k= \frac{1}{\sum_{i} q^{i}_k}\sum_{i} q^{i}_k(f_{i} - w_k)/\sigma_k,
\label{eq2}
\end{aligned}
\end{equation} 
where $\sigma_k$ is the column vector of $\Sigma$, $v'_k$ is a weighted average of the residuals between feature vector $f_i$ and $w_k$. $v_k$ means the representation for the $k$-th node, and forms the $k$-th column of the node feature matrix $\mathcal V$. $q_k^{i}$ is the soft-assignment of a feature vector $f_i$ to $w_k$, and can be computed by the following equation:
\begin{equation}  \textstyle
\begin{aligned}
\small
q_k^{i} = \frac{\exp(-||(f_i - w_k)/\sigma_k||_2^2/2)}{\sum_{j}\exp(-||(f_i - w_j)/\sigma_j||_2^2/2)},
\label{eq3}
\end{aligned}
\end{equation} 
where `$/$' means the element-wise division. Here, we compute the graph adjacent matrix by measuring the affinity between intra-node representations: $\mathcal A^{\tt intra} = f_{\tt norm}(\mathcal{V}^\mathrm{T} \times \mathcal{V}) \in \mathbb{\bm R}^{K\times K}$, where $f_{\tt norm}$ means the normalization operation.

\noindent {\bf \small \emph{{(2)} Cross-Graph Interaction}}$f_{\tt CGI}$. $f_{\tt CGI}$ models the between-graph interaction and guides the inter-graph message passing from $\mathcal V_C$ to $\mathcal V_E$. This goal leads us to draw inspiration from the non-local operation~\cite{wang2018non}, and compute inter-graph dependencies with attention mechanism. To begin with, as shown in Figure~\ref{fig:3}, we use different multi-layer perceptrons (MLPs)~\cite{qi2017pointnet} to transform $\mathcal V_C$ to the \emph{key} graph ${\mathcal V_C^\theta}$ and the \emph{value} graph $\mathcal V_C^\gamma$, and $\mathcal V_E$ to the \emph{query} graph $\mathcal V_E^\kappa$. Then, the similarity matrix $\mathcal A^{\tt inter}_{C\rightarrow E} \in \mathbb{\bm R}^{K\times K}$ is calculated by a matrix multiplication as:
\begin{equation}  \textstyle
\begin{aligned}
\small
\mathcal A^{\tt inter}_{C\rightarrow E} =  f_{\tt norm}({{\mathcal V_E^\kappa}^\mathrm{T} \times \mathcal V_C^\theta)},
\label{eq4}
\end{aligned}
\end{equation} 
where $\mathcal A^{\tt inter}_{C\rightarrow E}  \in \mathbb{\bm R}^{K\times K}$. After that, we can transfer semantic information from  $\mathcal V_C$ to $\mathcal V_E$ by
\begin{equation}  \textstyle
\begin{aligned}
\small
\mathcal V'_E = f_{\tt CGI} (\mathcal V_C, \mathcal V_E) = \chi(\mathcal A^{\tt inter}_{C\rightarrow E} \times \mathcal {V_C^\gamma}^\mathrm{T}) + \mathcal V_E,
\label{eq5}
\end{aligned}
\end{equation} 
where $\chi$ acts as the weighting parameter to adjust the importance of CGI \emph{w.r.t.} $\mathcal V_E$.

\noindent {\bf \small \emph{{(3)} Graph Reasoning}} $f_{\tt GR}$.
After performing inter-graph interaction, we conduct the intra-graph reasoning by taking ${\mathcal V}_C $ and $\mathcal V'_E$ as inputs to obtain enhanced graph representations. Here, $f_{\tt GR}$ can be implemented with graph convolution~\cite{kipf2016semi}:
\begin{equation}  \textstyle
\begin{aligned}
\small
\left\{ \begin{array}{ll} 
\mathbf V_C = f_{\tt GR} (\mathcal V_C) = {g}(\mathcal A_C^{\tt intra}\mathcal V_C W_C) \in \mathbb{\bm R}^{C\times K},\\
\mathbf V'_E = f_{\tt GR} (\mathcal V'_E) = {g}(\mathcal A_E^{\tt intra}\mathcal V'_E W_E) \in \mathbb{\bm R}^{C\times K},
&\end{array} 
\right. 
\label{eq6}
\end{aligned}
\end{equation} 
where $g(\cdot)$ is a non-linear activation function, $W_C$ and $W_E$ are learnable parameters of the graph convolution layer, and $\mathcal A_C^{\tt intra}$ and $\mathcal A_E^{\tt intra}$ denote the graph adjacent matrices for $\mathcal V_C$ and $\mathcal V'_E$, respectively. 
\begin{figure}[pt]
	\begin{center}
		\includegraphics[width=0.99\linewidth]{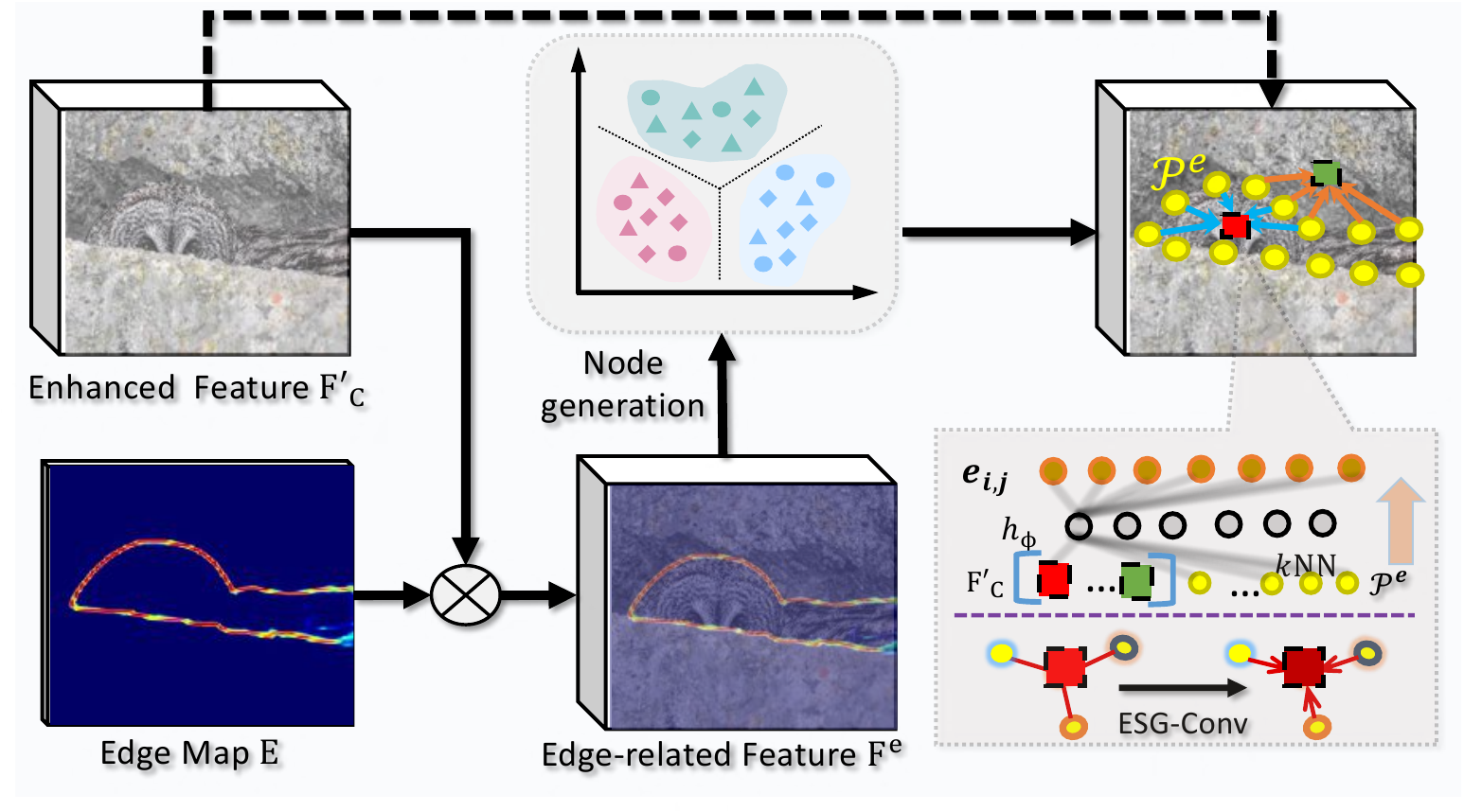}
	\end{center}
	\vspace{-20pt}
	\caption{{\bf Illustration of {ECGR}}. ECGR mines useful information from COEE to guide the representation learning of COD. The supportive nodes are generated in a `soft' way.}
	\label{fig:4e}
	\vspace{-1.em}
\end{figure}

\noindent {\bf \small \emph{{(4)} Graph Reprojection}}  $f_{\tt Rproj}$. 
To map the enhanced graph representations back to the original coordinate space, we revisit the assignments from the graph projection step. Formally, let us denote the assignment matrix for COD as $\mathcal Q_C = [{q_C}_k]_{k=0}^{(K-1)}$, where ${q_C}_k = [{q_C}_k^i]_{i=0}^{(h\times w)-1}$, and the assignment matrix for COEE as $\mathcal Q_E = [{q_E}_k]_{k=0}^{(K-1)}$, where ${q_E}_k = [{q_E}_k^i]_{i=0}^{(h\times w)-1}$. The graph reprojection $f_{\tt Rproj}$ can be formulated as:
\begin{equation}  \textstyle
\begin{aligned}
\small
\left\{ \begin{array}{ll} 
\hat{\mathbf F}_C &= \mathcal Q_C\mathbf V_C^\mathrm{T} + \mathbf F^l_C,~~\mathcal Q_C\in \mathbb{\bm R}^{(h\times w)\times K},\\
\breve{\mathbf F}_E &= \mathcal Q_E{\mathbf V'}_E^\mathrm{T} + \mathbf F^l_E,~~\mathcal Q_E \in \mathbb{\bm R}^{(h\times w)\times K},
\end{array} 
\right.
\label{eq7}
\end{aligned}
\vspace{-0.5em} 
\end{equation}
where $\hat{\mathbf F}_C \in \mathbb{\bm R}^{(h\times w)\times C}$ and $\breve{\mathbf F}_E \in \mathbb{\bm R}^{(h\times w)\times C}$ are the enhanced feature maps for COD and COEE respectively.

\noindent {\bf Edge-Constricted Graph Reasoning (ECGR).} 
ECGR focuses on edge-constricted relation reasoning in order to extract useful information from COEE to further guide the representation learning for COD. The idea illustration for our ECGR is given in Figure~\ref{fig:4e}.

\noindent {\bf \small \emph{{(1)} Our Goal}}. 
The goal of ECGR is to equip the model with an explicit edge perception capability so as to locate objects accurately. We expect $\hat{\mathbf F}_{C}$ to be updated by explicitly perceiving and encoding information about edge. With this goal, we first produce the enhanced feature map $\mathbf F'_C$ for COD by directly fusing $\breve{\mathbf F}_E$ and $\hat{\mathbf F}_{C}$ (via {\tt concatenate}), and then use a novel Edge Supportive Graph Convolution ({\tt ESG-Conv}) to update it, conditioned on $\mathbf E$. Next, we describe the edge supportive graph $\mathcal G^e(\mathbf E)$ and the graph convolution {\tt ESG-Conv} intended for it.

\noindent {\bf \small \emph{{(2)} Supportive Node/Vertex Generation}}. 
The first step for building $\mathcal G^e(\mathbf E)$ is to generate edge-based node embeddings. First, we map $\hat{\mathbf F}_E$ to a camouflage object-aware edge map $\mathbf E \in \mathbb{\bm R}^{h\times w\times 1}$ via a fully connected layer. Then, as shown in Figure 4, we obtain the edge-related features on regular grids of $\mathbf F'_C$ in a `soft' manner with the attention mechanism: $\mathbf F^e = \mathbf E \otimes \mathbf F'_C$, where $\otimes$ means the channel-wise multiplication operation. Finally, a graph projection operation $f_{\tt Gproj}$ is used to transform $\mathbf F^e$ into $z$ edge-based node embeddings, denoted as $\mathcal P^e = \{p^e_1, \cdots, p^e_z \}$, to represent the edge prior.

\noindent {\bf \small \emph{{(3)} Edge Supportive Graph Convolution {\tt ESG-Conv}}}. 
We construct our edge supportive graph $\mathcal G^{e}(\mathbf E) = (\mathcal V^{e}, \mathcal E^{e})$ as the k-nearest neighbor ($k$-NN) graph~\cite{wang2019dynamic} to link ${\mathbf F'}_{C}$ with $\mathcal P^e$, where $\mathcal V^{e}$ and $\mathcal E^{e}$ denote the vertices and edges respectively. Formally, we regard each feature vector ${f'^{c}_{i}} \in {\mathbf F'}_{C}$ as the central node and $\{p_j^c: (i,j)\in \mathcal E^{e}\}$ as its edge supportive nodes. The edge embedding $\mathbf e_{i,j}$ can be defined as:
\begin{equation}  \textstyle
\begin{aligned}
\small
\mathbf e_{i,j} = h_{\phi} ({f'^{c}_{i}}, p_j^c) = f_{\mathtt {Conv}} ({f'^{c}_{i}} - p_j^c),
\label{eq8}
\end{aligned}
\end{equation} 
where $h_{\phi}$ is a nonlinear function with learnable parameters $\phi$. The output of {\tt ESG-Conv} for the $i$-th feature vector/vertex is thus given as:
\begin{equation}  \textstyle
\begin{aligned}
\small
\breve{\mathbf f}_{i} = \max_{j: (i, j) \in \mathcal E^{e}} h_{\Phi} ({f'^{c}_{i}}, \mathbf e_{i,j}),
\label{eq9}
\end{aligned}
\end{equation} 
where $h_{\Phi}$ denotes the function for learning node embeddings with learnable parameters $\Phi$, and $\breve{\mathbf f}_i \in \breve{\mathbf F}_C$ means the evolved representation. With our {\tt ESG-Conv}, edge information can be explicitly encoded into underlying representations, \emph{i.e.,} $\breve{\mathbf F}_C = \mathtt{ESGConv} (\mathbf F'_C; \mathcal G^e(\mathbf E))$.

\noindent {\bf Recurrent Learning Process.} 
To fully exploit the mutual benefits between COD and COEE, we can further formulate our {MGL} as the following recurrent learning process:
\begin{equation}  \textstyle
\begin{aligned}
\small
\left\{ \begin{array}{ll} 
&\breve{\mathbf F}^{(t+1)}_E = f_{\tt RIGR} (\breve{\mathbf F}^{(t)}_C, \breve{\mathbf F}^{(t)}_E), \\ 
&\breve{\mathbf F}^{(t+1)}_C = f_{\tt ECGR} (\breve{\mathbf F}^{(t)}_C, \breve{\mathbf F}^{(t+1)}_E, \mathbf E^{(t+1)}),
\end{array} 
\right.
\label{eq10}
\end{aligned}
\end{equation} 
where $f_{\tt RIGR}$ and $f_{\tt ECGR}$ means {\bf RIGR} and {\bf ECGR} modules respectively. Note that at the beginning ($t=1$), $\breve{\mathbf F}^{(1)}_C = f_{\tt MTFE} (\mathbf I; \Theta_{\tt MTFE})$ and $\breve{\mathbf F}^{(1)}_E = f_{\tt MTFE} (\mathbf I; \Theta_{\tt MTFE})$.

\begin{table*}[t!]
	\centering
	\caption{{\bf Quantitative results on different datasets}. `$\dagger$' means SOTA methods for GOD and SOD. $\uparrow$ (or $\downarrow$) indicates that the higher (or the lower) the better. 
	Online benchmark: \url{http://dpfan.net/camouflage}.
	}\label{tab:1}
	\vspace{-10pt}
	\scriptsize
	\renewcommand{\arraystretch}{1.0}
	\setlength\tabcolsep{8pt}
		\begin{tabular}{l|cccc||cccc||cccc}
			\hline
			\toprule
			& \multicolumn{4}{c||}{\tabincell{c}{CHAMELEON~\cite{skurowski2018animal}}} 
			&\multicolumn{4}{c||}{\tabincell{c}{CAMO-Test~\cite{ltnghia-CVIU2019}}} 
			& \multicolumn{4}{c}{ \tabincell{c}{COD10K-Test~\cite{fan2020camouflaged}}} \\
			\cline{2-13}
			Methods~~~ &$S_\alpha\uparrow$      &$E_\phi\uparrow$     &$F_\beta^w\uparrow$      &$M\downarrow$
			&$S_\alpha\uparrow$      &$E_\phi\uparrow$     &$F_\beta^w\uparrow$      &$M\downarrow$
			&$S_\alpha\uparrow$      &$E_\phi\uparrow$     &$F_\beta^w\uparrow$      &$M\downarrow$ \\
			\hline
			2017~FPN~$\dagger$~\cite{lin2017feature}
			&0.794&0.783&0.590&0.075&0.684&0.677&0.483&0.131&0.697&0.691&0.411&0.075\\
			2017~MaskRCNN~$\dagger$~\cite{he2017mask}
			&0.643&0.778&0.518&0.099&0.574&0.715&0.430&0.151&0.613&0.748&0.402&0.080\\
			2017~PSPNet~$\dagger$~\cite{zhao2017pyramid}
			&0.773&0.758&0.555&0.085&0.663&0.659&0.455&0.139&0.678&0.680&0.377&0.080\\
			2018~UNet++~$\dagger$~\cite{zou2018DLMIA}
			&0.695&0.762&0.501&0.094&0.599&0.653&0.392&0.149&0.623&0.672&0.350&0.086\\
			2018~PiCANet~$\dagger$~\cite{liu2018picanet}
			&0.769&0.749&0.536&0.085&0.609&0.584&0.356&0.156&0.649&0.643&0.322&0.090\\
			2019~MSRCNN~$\dagger$~\cite{huang2019mask}
			&0.637&0.686&0.443&0.091&0.617&0.669&0.454&0.133&0.641&0.706&0.419&0.073\\
			2019~PoolNet~$\dagger$~\cite{liu2019simple}
			&0.776&0.779&0.555&0.081&0.702&0.698&0.494&0.129&0.705&0.713&0.416&0.074\\
			2019~BASNet~$\dagger$~\cite{qin2019basnet}
			&0.687&0.721&0.474&0.118&0.618&0.661&0.413&0.159&0.634&0.678&0.365&0.105\\ 
			2019~PFANet~$\dagger$~\cite{zhao2019pyramid}
			&0.679&0.648&0.378&0.144&0.659&0.622&0.391&0.172&0.636&0.618&0.286&0.128\\
			2019~CPD~$\dagger$~\cite{wu2019cascaded}
			&0.853&0.866&0.706&0.052&0.726&0.729&0.550&0.115&0.747&0.770&0.508&0.059\\ 
			2019~HTC~$\dagger$~\cite{chen2019hybrid}
			&0.517&0.489&0.204&0.129&0.476&0.442&0.174&0.172&0.548&0.520&0.221&0.088\\
			2019~EGNet~$\dagger$~\cite{zhao2019egnet}
			&0.848&0.870&0.702&0.050&0.732&0.768&0.583&0.104&0.737&0.779&0.509&0.056\\
			2019~ANet-SRM~\cite{ltnghia-CVIU2019}
			& $\ddagger$ & $\ddagger$ & $\ddagger$& $\ddagger$  &0.682&0.685&0.484&0.126& $\ddagger$ & $\ddagger$ & $\ddagger$ & $\ddagger$\\
			2020~MirrorNet~\cite{yan2020}
			& $\ddagger$ & $\ddagger$ & $\ddagger$& $\ddagger$  &0.741&0.804&0.652&0.100& $\ddagger$ & $\ddagger$ & $\ddagger$ & $\ddagger$\\
			2020~PraNet~\cite{fan2020pranet} 
			&{0.860}&{0.898}&{0.763}&{0.044}&{0.769}&{0.833}&{0.663}&{0.094}&{0.789}&{0.839}&{0.629}&{0.045}\\
			2020~SINet~\cite{fan2020camouflaged} 
			&{0.869}&{0.891}&{0.740}&{0.044}&{0.751}&{0.771}&{0.606}&{0.100}&{0.771}&{0.806}&{0.551}&{0.051}\\
			\hline
			{\bf S-MGL} (ours)
			&{0.892}&{0.921}&{0.803}&{0.032}&{0.772}&\textbf{0.850}&{0.664}&{0.089}&{0.811}&{0.851}&{0.655}&{0.037} \\
			{\bf R-MGL} (ours) 
			&\textbf{0.893}&\textbf{0.923}&\textbf{0.813}&\textbf{0.030}&\textbf{0.775}&{0.847}&\textbf{0.673}&\textbf{0.088}&\textbf{0.814}&\textbf{0.865}&\textbf{0.666}&\textbf{0.035} \\		
			\hline
			\toprule
		\end{tabular}
	\vspace{-10pt}
\end{table*}

\subsection{Implementation Details}
We present two versions of MGL. One, named as {\bf S-MGL}, is a single-stage model which mines the mutual information only once. The other, named as {\bf R-MGL}, includes a recurrent learning process performing two recurrent stages. The implementation is detailed as follows:


\noindent {\bf Multi-Task Feature Extractor.}
Following existing arts~\cite{fan2020camouflaged}, we employ ResNet-50~\cite{He_2016_CVPR} pre-trained on ImageNet~\cite{krizhevsky2012imagenet} as the backbone. We use the dilated network technique~\cite{yu2015multi} to ensure that the feature map for COD ($\mathbf F_C$) is $60\times 60$ in resolution. To extract features for COEE ($\mathbf F_E$), we first collect a set of side-output features $\{\mathbf S_k\}_{k=2}^5$ from ResNet-50, then make these features have the same resolution of $60\times60$ via a bi-linear up/down-sampling layer, and finally fuse them with a {\tt concatenate} layer followed by a $1\times1$ convolutional layer.

\noindent {\bf Region-Induced Graph Reasoning Module.} 
We follow ~\cite{li2018beyond} to design and implement $f_{\tt Gproj}$, and encode $\mathbf F_C$ and $\mathbf F_E$ to $K=32$ semantic nodes respectively (see Table~\ref{tab:5}). The transformation function in $f_{\tt CGI}$ is implemented by MLPs ($1\times1$ convolution). In our RIGR, Eq.~\ref{eq4} is used to build the \emph{between-graph} relations, and Eq.~\ref{eq5} is used to capture semantic guidance information (from $\mathcal V_C$ to $\mathcal V_E$) and produce the evolved graph representation $\mathbf V'_E$ for $\mathcal V'_E$. $f_{\tt GR}$ is implemented via GCNs~\cite{kipf2016semi} and $f_{\tt Rproj}$ reuses the assignment matrix for graph re-projection by using Eq.~\ref{eq7}.

\noindent {\bf Edge-Constricted Graph Reasoning Module.} 
For the number of edge supportive nodes, we observe that $z = 32$ can ensure a promising speed-accuracy tradeoff (see Table~\ref{tab:5}). $h_{\phi}(\cdot)$ in Eq.~\ref{eq8} can be simply implemented with element-wise \emph{subtraction} operation followed by a $1\times1$ convolution. $h_{\Phi}(\cdot)$ in Eq.~\ref{eq9} concatenates edge and node embeddings, \emph{i.e.,} ${f'^{c}_{i}}$ \& $\mathbf e_{i,j}$, and uses a $1\times1$ convolution to fuse them for producing $\breve{\mathbf f}_i \in \breve{\mathbf F}_C$.

\begin{table*}[t!]
	\centering
	\vspace*{-8pt}
\caption{{\bf Ablation study} of the proposed approach on CHAMELEON, CAMO {\tt test} and COD10K {\tt test}.}\label{tab:3}
	\vspace*{-8pt}
	\scriptsize
	\renewcommand{\arraystretch}{1.0}
	\setlength\tabcolsep{6.3pt}
		\begin{tabular}{cccc|cccc||cccc||cccc}
			\hline
			\toprule
			\multicolumn{4}{c|}{\tabincell{c}{Candidate}}  & \multicolumn{4}{c||}{\tabincell{c}{CHAMELEON~\cite{skurowski2018animal}}} 
			&\multicolumn{4}{c||}{\tabincell{c}{CAMO-Test~\cite{ltnghia-CVIU2019}}} 
			& \multicolumn{4}{c}{ \tabincell{c}{COD10K-Test~\cite{fan2020camouflaged}}} \\
			\cline{1-16}
			
			ResNet-50 &RIGR     &ECGR     &RL
			&$S_\alpha\uparrow$      &$E_\phi\uparrow$     &$F_\beta^w\uparrow$      &$M\downarrow$
			&$S_\alpha\uparrow$      &$E_\phi\uparrow$     &$F_\beta^w\uparrow$      &$M\downarrow$
			&$S_\alpha\uparrow$      &$E_\phi\uparrow$     &$F_\beta^w\uparrow$      &$M\downarrow$ \\
			\hline
			\ding{52}&&&&0.767&0.799&0.535&0.094&0.742&0.786&0.538&0.130&0.729&0.692&0.436&0.079\\
			\ding{52}&\ding{52}&&&0.844&0.863&0.686&0.055&0.766&0.828&0.611&0.104&0.785&0.758&0.557&0.052\\
			\ding{52}&\ding{52}&\ding{52}&&{0.892}&{0.921}&{0.803}&{0.032}&{0.772}&\textbf{0.850}&{0.664}&{0.089}&{0.811}&{0.851}&{0.655}&{0.037} \\
			\ding{52}&\ding{52}&\ding{52}&\ding{52}				&\textbf{0.893}&\textbf{0.923}&\textbf{0.813}&\textbf{0.030}&\textbf{0.775}&{0.847}&\textbf{0.673}&\textbf{0.088}&\textbf{0.814}&\textbf{0.865}&\textbf{0.666}&\textbf{0.035} \\
			\toprule
		\end{tabular}
	\vspace{-5pt}
\end{table*}

\noindent {\bf Classifier and Loss Function.} 
After obtaining the evolved representations $\breve{\mathbf F}^{(t)}_E$ and $\breve{\mathbf F}^{(t)}_C$, we use classifiers to map them to the corresponding outputs $\mathbf E$ and $\mathbf C$, which are implemented by $1\times1$ convolutional layers. For training, we use bi-linear interpolation to upsample the output maps to the original size to calculate the loss. We use the cross-entropy loss~\cite{de2005tutorial} for both tasks:
\begin{equation}  \textstyle
\begin{aligned}
\small
L = L^c_{CE} (\mathbf C, \mathbf G_C) + \gamma L^e_{CE} (\mathbf E, \mathbf G_E),
\label{eq11}
\end{aligned}
\end{equation} 
where $\mathbf G_C$ and $\mathbf G_E$ mean the groundtruth labels, and $\gamma$ means the combination weight. Here we simply set $\gamma =1$.

\section{Experiments}
\subsection{Experimental Setup}
\noindent {\bf Datasets:}
We perform extensive experiments on the following public benchmarks:
\begin{itemize}[leftmargin=*]
	\vspace{-0.6em}
	\item {\bf CHAMELEON~\cite{skurowski2018animal}} includes $76$ high-resolution images finely annotated with pixel-level labels. All images in CHAMELEON are collected from the Internet.
	
	\vspace{-0.6em}
	\item {\bf CAMO~\cite{ltnghia-CVIU2019}} is a collection of $2,500$ images with $8$ categories. In this dataset, both naturally camouflaged objects and artificially camouflaged objects are collected with finely-annotated labels.
	
	\vspace{-0.6em}
	\item {\bf COD10K~\cite{fan2020camouflaged}} is the largest COD dataset, which includes $10,000$ images with $10$ super-classes and $78$ sub-classes. All images are collected from photography websites.
	\vspace{-0.6em}
\end{itemize}
Our {\tt train set} is a combination of the {train} sets from CAMO and COD10K provided by~\cite{fan2021cancealed}.

\noindent {\bf Evaluation Metric:}
Following~\cite{fan2020camouflaged,ltnghia-CVIU2019}, we adopt mean absolute error (MAE) as evaluation metric. In addition, mean E-measure ($E_\phi$)~\cite{fan2018enhanced},  S-measure ($S_\alpha$)~\cite{fan2017structure} and weighted F-measure ($F^w_\beta$)~\cite{margolin2014evaluate} are used for balanced comparisons. Moreover, for evaluating our auxiliary COEE task, we adopt the precision-recall metric with F-measure following~\cite{xie2015holistically}. 
Evaluation tools: \url{https://github.com/DengPingFan/CODToolbox}.

\noindent {\bf Training Settings:}
During training, the weights of MTFE are initialized by ResNet-50~\cite{He_2016_CVPR} pre-trained on ImageNet~\cite{krizhevsky2012imagenet}, and the remaining layers/modules are randomly initialized. For data preparation, we perform data augmentation techniques on all training data, including random cropping, left-right flipping and scaling in the range of $[0.75,1.25]$. For optimization, we use the Stochastic Gradient Descent (SGD) with `poly' learning rate scheduling policy: $lr = base\_{lr} \times (1- \frac{iter}{max_{iter}})^{power}$.  The base learning rate $base\_{lr}$ is set to $10^{-7}$ and $power$ to $0.9$.

\noindent {\bf Reproducibility:} 
Our {S-MGL} and {R-MGL} are implemented based on PyTorch. Our model is trained on a NVIDIA Tesla V100 GPU to ensure a larger batch size. During test, all models are performed on a NVIDIA GTX Titan X GPU with 12G memory.

\subsection{Comparison with State-of-the-Arts}
\noindent {\bf Baselines / SOTAs:} 
Similar to~\cite{fan2020camouflaged}, we first select strong baseline models which achieve SOTA performance in closely related fields, \emph{i.e,} GOD and SOD. Moreover, all recently published methods for COD are included for comparisons. In sum, we compare our methods (S-MGL and R-MGL) against $16$ SOTAs, which are trained under their recommended settings with the same {\tt train} set as ours. 

\noindent {\bf Performance on CHAMELEON:} 
Table~\ref{tab:1} reports the comparison results with $14$ SOTAs on CHAMELEON. For fair comparison, all models use the same {\tt train} set for training. As can be seen, our S-MGL achieves better performance than all compared works across all metrics. When compared with the state-of-the-art SINet~\cite{fan2020camouflaged}, S-MGL significantly lowers MAE by $27.3\%$ and improve $F^w_\beta$ by $8.5\%$. Our R-MGL further boosts the performance and sets a new record. Clearly, our solution can significantly overcome the ambiguity in camouflaged scenes and provide more reliable results than existing approaches.

\noindent {\bf Performance on CAMO:}
We also compare our methods with SOTAs on CAMO {\tt test}. As can be seen in Table~\ref{tab:1}, our S-MGL and R-MGL achieve significantly better performance than other solutions. This is because our model can fully exploit mutual benefits and ensure model's reliability to overcome the heavy occlusions and indefinable boundaries in complex scenes.

\begin{table}[t!]
	\centering
\vspace*{-8pt}
\caption{{\bf Quantitative results} of different underlying feature enhancement algorithms.}\label{tab:4}
\vspace*{-8pt}
	\scriptsize
	\renewcommand{\arraystretch}{1.0}
	\setlength\tabcolsep{1.4pt}
		\begin{tabular}{ll|cccc|cccc}
			\hline
			\toprule
			& 
			&\multicolumn{4}{c|}{\tabincell{c}{CAMO-Test~\cite{ltnghia-CVIU2019}}} 
			& \multicolumn{4}{c}{ \tabincell{c}{COD10K-Test~\cite{fan2020camouflaged}}} \\
			\cline{3-10}
			Method&
			&$S_\alpha\uparrow$      &$E_\phi\uparrow$     &$F_\beta^w\uparrow$      &$M\downarrow$
			&$S_\alpha\uparrow$      &$E_\phi\uparrow$     &$F_\beta^w\uparrow$      &$M\downarrow$ \\
			\hline
			Baseline (ResNet-50 FCN)&&0.742&0.786&0.538&0.130&0.729&0.692&0.436&0.079\\
			Baseline + NL~\cite{wang2018non}&&0.748&0.791&0.541&0.122&0.731&0.711&0.459&0.073\\
			MTFE + MUL~\cite{nie2018mutual} &&0.751&0.799&0.551&0.118&0.736&0.721&0.498&0.070\\
			{\bf S-MGL} (ours)&&{0.772}&\textbf{0.850}&{0.664}&{0.089}&{0.811}&{0.851}&{0.655}&{0.037}\\
			{\bf R-MGL} (ours)&&\textbf{0.775}&{0.847}&\textbf{0.673}&\textbf{0.088}&\textbf{0.814}&\textbf{0.865}&\textbf{0.666}&\textbf{0.035} \\
			\toprule
		\end{tabular}
	\vspace{-10pt}	
\end{table}

\begin{figure*}[pt]
	\begin{center}
		\includegraphics[width=0.999\linewidth]{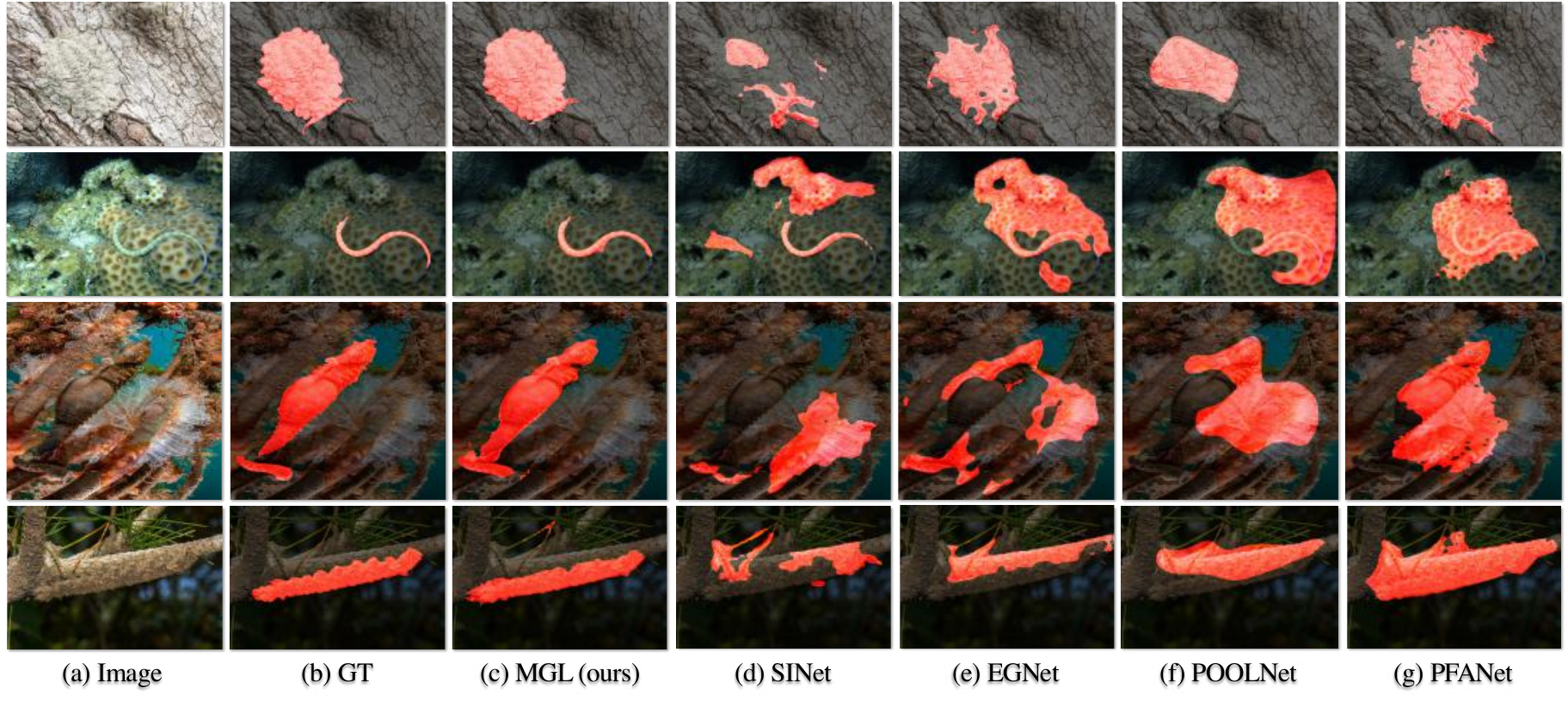}
	\end{center}
	\vspace{-25pt}
	\caption{{\bf Qualitative comparisons} between different models: (c) our approach (R-MGL), (d) SINet~\cite{fan2020camouflaged}, (e) EGNet~\cite{zhao2019egnet}, (f) POOLNet~\cite{liu2019simple}, and (g) PFANet~\cite{zhao2019pyramid}. Clearly, our approach can better spot hidden objects with more clear boundaries.}
	\label{fig:visual}
	\vspace{-10pt}
\end{figure*}

\noindent {\bf Performance on COD10K:}
On the largest COD10K {\tt test}, our solution sets new records for all metrics. Specifically, S-MGL greatly surpasses currently best models, which achieves $S_\alpha$ score of $81.1\%$, $E_\phi$ score of $85.1\%$, $F_\beta^w$ score of $65.5\%$, and sets the best MAE score of $0.037$. R-MGL further boosts the performance. The powerful graph-based interaction modules enable our models to work well with the auxiliary COEE for overcoming all challenges in COD. Some visual samples are given in Figure~\ref{fig:visual}.

\begin{figure}[pt]
	\begin{center}
		\includegraphics[width=0.99\linewidth]{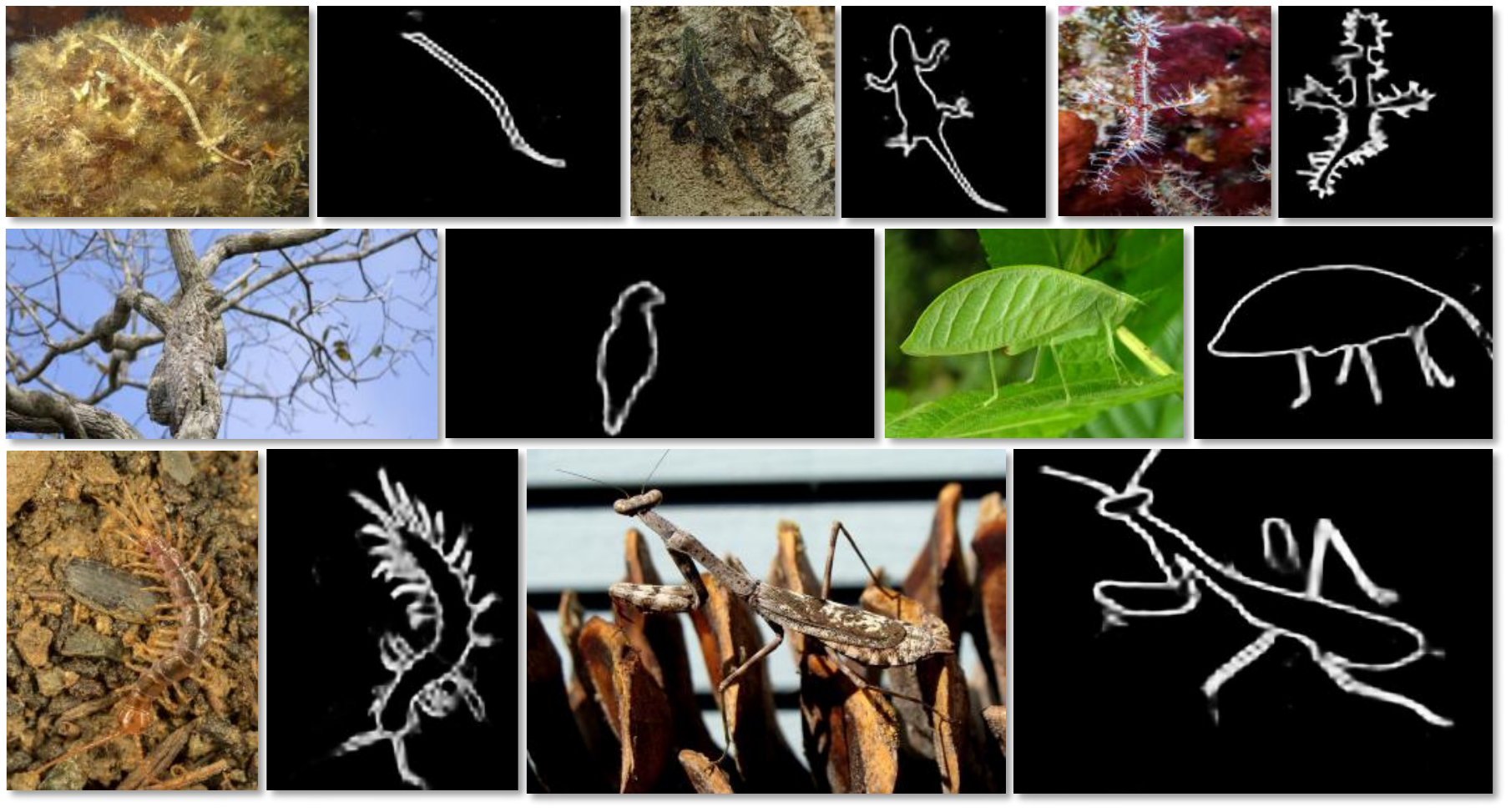}
	\end{center}
	\vspace{-20pt}
	\caption{{{\bf Visual results} of our approach (R-MGL) for camouflaged object-aware edge extraction} on COD10K {\tt test}. }
	\label{fig:PC}
	\vspace{-1em}	
\end{figure}

\noindent {\bf Auxiliary Task (COEE):} We believe that the mutual learning within our model can also significantly benefit the auxiliary COEE. To verify this, we compare our MGL with the well-known HED~\cite{xie2015holistically} and its improved version DSS~\cite{hou2017deeply}. Moreover, we include the strong multi-task baseline MUL~\cite{nie2018mutual} for comparison. All models are trained on the same {\tt train} set with our extracted edge labels. As shown in Table~\ref{tab:coee}, our S-MGL and R-MGL achieve stronger results than existing models in this task, which shows that our solution can not only improve the performance of the main task (COD) but also boost the auxiliary task (COEE). Some visual samples are provided in Figure~\ref{fig:PC}.

\subsection{Ablation Study}

\noindent {\bf Effectiveness of RIGR and ECGR:}
To verify the effect of our RIGR, we use a model based on ResNet50-FCN as the baseline. First, as shown in Table~\ref{tab:3}, RIGR enables the model to achieve a certain performance improvement compared to the baseline across all datasets, which demonstrates the effectiveness of the proposed RIGR. Besides, by adding ECGR, we can see a further improvement in accuracy. Thus, it is clear that improving the true edge visibility is important and can empower the model with stronger capability for overcoming difficulties in COD tasks. Moreover, we have carefully studied the parameters in our RIGR and ECGR modules. Table~\ref{tab:5} provides the detailed comparisons of different settings. 

\begin{table}[t!]
	\centering
	\vspace*{-8pt}
\caption{{\bf Detailed ablation study} of different parameter settings. `K' means the number of semantic nodes; `z' stands for the number of edge supportive nodes; `t' means that t recurrent stages are used in our MGL.}\label{tab:5}
	\vspace*{-8pt}
	\scriptsize
	\renewcommand{\arraystretch}{1.0}
	\setlength\tabcolsep{1.0pt}
		\begin{tabular}{ll|cccc|cccc}
			\hline
			\toprule
			& 
			&\multicolumn{4}{c|}{\tabincell{c}{CAMO-Test~\cite{ltnghia-CVIU2019}}} 
			& \multicolumn{4}{c}{ \tabincell{c}{COD10K-Test~\cite{fan2020camouflaged}}} \\
			\cline{3-10}
			Method&
			&$S_\alpha\uparrow$      &$E_\phi\uparrow$     &$F_\beta^w\uparrow$      &$M\downarrow$
			&$S_\alpha\uparrow$      &$E_\phi\uparrow$     &$F_\beta^w\uparrow$      &$M\downarrow$ \\
			\hline
			{\bf S-MGL} (K=16, z=32)&&0.771&0.832&0.661&0.092&0.805&0.832&0.638&0.042\\
			{\bf S-MGL} (K=32, z=32)&&{0.772}&\textbf{0.850}&{0.664}&{0.089}&{0.811}&{0.851}&{0.655}&{0.037}\\
			{\bf S-MGL} (K=64, z=32)&&0.774&0.849&0.661&0.089&0.809&0.854&0.648&0.037\\
			\hline
			{\bf S-MGL} (K=32, z=16)&&0.772&0.843&0.662&0.090&0.804&0.837&0.640&0.040\\
			{\bf S-MGL} (K=32, z=32)&&{0.772}&\textbf{0.850}&{0.664}&{0.089}&{0.811}&{0.851}&{0.655}&{0.037}\\
			{\bf S-MGL} (K=32, z=64)&&0.773&0.848&0.666&0.089&0.807&0.855&0.657&0.037\\
			\hline
			{\bf R-MGL} (K=32, z=32, t=1)&&{0.772}&\textbf{0.850}&{0.664}&{0.089}&{0.811}&{0.851}&{0.655}&{0.037}\\
			{\bf R-MGL} (K=32, z=32, t=2)&&\textbf{0.775}&{0.847}&\textbf{0.673}&\textbf{0.088}&{0.814}&\textbf{0.865}&\textbf{0.666}&\textbf{0.035} \\
			{\bf R-MGL} (K=32, z=32, t=3)&&0.773&0.848&0.672&\textbf{0.088}&\textbf{0.815}&0.862&0.661&0.036\\
			\toprule
		\end{tabular}
	\vspace{-10pt}	
\end{table}

\noindent {\bf Usefulness of Recurrent Learning:} 
We can easily extend our MGL into a more comprehensive recurrent reasoning process. Table~\ref{tab:3} shows that model's performance can be further improved with recurrent learning techniques. This is because the recurrent process can be used to refine the initial results / features, and thus improve the accuracy. Furthermore, according to our experiments (see Table~\ref{tab:5}), using only two recurrent steps can ensure promising performance, which makes our R-MGL set new records for all benchmarks and greatly outperform existing approaches. 

\begin{table}[pt!]
	\centering
\caption{{\bf The comparison of camouflaged object-aware edge results} with some wide-used methods on CAMO {\tt test} and COD10K {\tt test}.}\label{tab:coee}
	\vspace*{-8pt}
	\scriptsize
	\renewcommand{\arraystretch}{1.0}
	\setlength\tabcolsep{10pt}
		\begin{tabular}{l|cc|cc}
			\hline
			\toprule
			
			&\multicolumn{2}{c|}{\tabincell{c}{CAMO-Test~\cite{ltnghia-CVIU2019}}} 
			& \multicolumn{2}{c}{ \tabincell{c}{COD10K-Test~\cite{fan2020camouflaged}}} \\
			\cline{2-5}
			Method
			&ODS     &OIS    &ODS     &OIS\\
			\hline
			HED~\cite{xie2015holistically} &0.315&0.318&0.294&0.313\\
			DSS~\cite{hou2017deeply} &0.316&0.336&0.347&0.372\\
			\hline
			Res50-FCN&0.509&0.511&0.505&0.524\\
			MTEF + MUL~\cite{nie2018mutual} &0.521&0.539&0.516&0.534\\
			\hline
			{\bf S-MGL}&0.536&0.545&0.535&0.557\\
			{\bf R-MGL}&\bf 0.543&\bf 0.551&\bf0.540&\bf0.558\\
			\toprule
		\end{tabular}
	\vspace{-20pt}	
\end{table}

\noindent {\bf Superiority of Mutual Graph Learning:}
We conduct comprehensive experiments / comparisons to show the superiority of our mutual graph learning approach. As shown in Table~\ref{tab:4}, compared with the widely used non-local (NL) operation, the explicit mutual learning (MUL) can guarantee more reliable results, which demonstrates that mining the valuable auxiliary edge information can help the model overcome COD challenges, such as heavy occlusions and indefinable boundaries. Our idea is to extend MUL from regular grids to graph domain. Clearly, our S-MGL and R-MGL outperform conventional MUL due to its stronger capability for capturing high-order relations. These experiments demonstrate that deeply mining high-order relations between COD and auxiliary COEE is meaningful, which can significantly improve the reliability of model to better overcome the intrinsic ambiguity for the challenging COD task. Moreover, reasoning high-order relations through graphs would bring 
clear performance improvements.

\section{Conclusion}
We have presented the Mutual Graph Learning (MGL), a graph-based, joint learning framework for detecting camouflaged objects and their true edges. Our model includes two novel neural modules: Region-Induced Graph Reasoning (RIGR) module and Edge-Constricted Graph Reasoning (ECGR) module, which can work together to mine valuable complementary information for improving the true edge visibility for COD. We also formulate our MGL as a recurrent graph reasoning process to fully exploit all useful information. Extensive experiments show that explicitly mining true edge prior / information can help to overcome the intrinsic difficulties in COD tasks, such as occlusions and indefinable boundaries. We believe our MGL can also benefit other related computer vision tasks, \emph{e.g.,} panoptic segmentation, that require multi-source information for the joint representation enhancement.

\noindent {\bf Acknowledgement.}
This research was funded in part by	the National Natural Science Foundation of China (U1964203) and the National Key R\&D Program Project of China (2017YFB0102603).

{\small
	\bibliographystyle{ieee_fullname}
	\bibliography{egbib}
}
\end{document}